# CARE: Coherent Actionable Recourse based on Sound Counterfactual Explanations

Peyman Rasouli
Department of Informatics
University of Oslo
Oslo, Norway
peymanra@ifi.uio.no

Ingrid Chieh Yu
Department of Informatics
University of Oslo
Oslo, Norway
ingridcy@ifi.uio.no

## ABSTRACT

Counterfactual explanation methods interpret the outputs of a machine learning model in the form of "what-if scenarios" without compromising the fidelity-interpretability trade-off. They explain how to obtain a desired prediction from the model by recommending small changes to the input features, aka recourse. We believe an actionable recourse should be created based on sound counterfactual explanations originating from the distribution of the ground-truth data and linked to the domain knowledge. Moreover, it needs to preserve the coherency between changed/unchanged features while satisfying user/domain constraints. This paper introduces CARE, a modular explanation framework that addresses the model- and user-level desiderata in a consecutive and structured manner. We tackle the existing requirements by proposing novel and efficient solutions that are formulated in a multi-objective optimization framework. The designed framework enables including arbitrary requirements and generating counterfactual explanations and actionable recourse by choice. As a model-agnostic approach, CARE generates multiple, diverse explanations for any black-box model in tabular classification and regression settings. Several experiments on standard data sets and black-box models demonstrate the effectiveness of our modular framework and its superior performance compared to the baselines.

## 1 INTRODUCTION

"Why was my loan application denied?" is an example of a follow-up question when a Machine Learning (ML) system does not provide a user's desired outcome. The user often expects more information beyond merely an explanation to know "What do I need to change for the bank to approve my loan?". Given the user's preferences, an actionable recourse is a list of possible changes for achieving the desired decision. By recommending the alterations, it also explains the reasons behind the current outcome. Counterfactual explanation is a technique for identifying the necessary changes in the original input that leads to the desired prediction of the ML model [24]. For example, a possible counterfactual for a user who is denied a loan by the model may be like this: "if your balance was 5000$, your loan application would have been accepted". This explanation justifies the denial prediction of the model, gives a guideline to reverse the prediction, reveals potential fairness issues, and demonstrates the correlations and patterns among the domain's data. For having an actionable and realistic recourse, the counterfactual explanation should be personalized with respect to the user and be sound in the sense that it originates from the domain's observations. We claim that the solution should satisfy several important desiderata in addition to basic properties such as feature value similarity and opposite prediction. In the following, we discuss these requirements in conjunction with the relevant existing studies.

*Proximity.* A counterfactual should provide plausible changes to the original input that are in accordance with the observations of the model. For example, "earning negative money for granting the loan" is an *outlier* counterfactual which is untrustworthy and unrealistic. This notion is called *proximity*, which indicates a counterfactual instance should lie in the neighborhood of the ground-truth data [12]. Recent papers have proposed various ways of handling *proximity*. DACE uses Local Outlier Factor [1] in the cost function of the optimization algorithm [10]. Several works employ generative models (e.g., Variational Auto-Encoders) for approximating the data distribution, and then sample realistic counterfactual instances [9, 15, 23]. FACE respects the underlying data distribution by connecting the counterfactual to the input via high-density paths [18]. MOC uses a weighted average distance metric between the original input and its $K$ nearest instances in the training data [4].

*Connectedness.* A counterfactual should be the result of existing knowledge and not a consequence of an artifact of the ML model. Artifacts can be created due to lack of training data for some regions in the feature space that diminish the robustness of the model and provoke misbehavior. Having an explanation caused by an artifact is not associated to the domain knowledge and is undesirable in the context of interpretability and feasibility [12]. Therefore, there is a need to define a relationship between a counterfactual instance and existing knowledge (training data) using a path. This notion is called *connectedness*, which implies that an interpretable counterfactual should be continuously connected to data points from the same class [12]. This property is thus complementary to *proximity*, as two instances can be close but not linked by a continuous path. It leads to interpretable counterfactuals that comply with the domain knowledge and, therefore, more actionable.

*Coherency.* A counterfactual should preserve the correlation among features to create a coherent explanation. For example, a counterfactual that provides the desired outcome via "increasing the education level from Bachelor to Master" without increasing "age" is not feasible because "age" and "education" are two correlated features, and they need to be changed jointly. This counterfactual may still satisfy soundness properties such as *proximity* and *connectedness* (as there can be found samples in the training data with the same age as the input but having a Master's degree), however, from the actionability point of view, it is an inconsistent explanation. Considering this property is even more essential when a user imposes constraints over some features regardless of the status of other features. In this case, the counterfactual needs to implicitly preserve the consistency of changed/unchanged features. Although there are several research works around creating statistically sound explanations [4, 9, 10, 15, 18, 23], to the best of our knowledge, the *coherency* property has been remained unexplored.

*Actionability.* A counterfactual should satisfy some global and local preferences that are domain-specific and defined by the end-user. Global preferences are constraints that should be satisfied by every counterfactual, for instance, fixing immutable features like **race** and **gender**, while local preferences are related to individual instances. For example, a possible range for feature **balance** for an individual is [3000$, 6000$] while for another one is [5000$, 10000$]. This notion is called *actionability*, implying a recourse that meets the user's preferences [22]. Recent works address the actionability property either by classifying the features into immutable and mutable types [4, 11, 17], or by defining a range/set of values for numerical and categorical features [21, 22].

According to our literature review on counterfactual explanation methods, the majority of works focus on *proximity* and *actionability* properties, while *connectedness* and *coherency* received little attention. A likely reason for neglecting connectedness as an objective goal in the counterfactual generation process can be its computational burden. Moreover, many state-of-the-art works use proximity as an interchangeable property for connectedness, which is arguable, as proximity prevents generating outlier counterfactuals while connectedness results in counterfactuals that are connected to the existing domain knowledge. Disregarding connectedness results in instances being created from areas where the model has no information about (artifacts) and makes questionable improvisations (decisions) [13]. In our opinion, both proximity and connectedness properties are necessary for deriving statistically sound counterfactual explanations.

Most of the stated works establish proximity by finding a counterfactual that is connected to the entire training data. Indeed, this can be problematic, because a sparse counterfactual instance (changes in one/two features) lies fairly close to the overall training data, however, it can be out-of-distribution with respect to the subset of data that share the same values as the changed features and belong to the same class [23]. In contrast, our definition of proximity is inline with [23] which states that a counterfactual instance should be an inlier with respect to a subset of data that are similar and belong to the same class as the counterfactual.

The mentioned statistical properties do not preserve the consistency among features, especially when user's preferences come into play. Often, a user defines constraints for some features irrespective of the status of others. When an algorithm only satisfies the specified constraints and neglect their effects on the other features, it will generate an unrealistic and impractical recourse. Hence, we view *coherency* and *actionability* as two intertwined properties that should be satisfied simultaneously.

In this paper, we propose a method for generating **C**oherent **A**ctionable **R**ecourse based on sound counterfactual **E**xplanations (**CARE**). CARE provides actionable recourse by fulfilling the mentioned desiderata through objective functions organized in a modular hierarchy structure and optimized using Non-dominated Sorting Genetic Algorithm III (NSGA-III) [5]. The optimization choice leads to a model-agnostic explanation method that generates multiple (diverse) counterfactuals for every input, applies to both classification and regression tasks, and handles mixed-feature data sets. We propose a novel approach to preserve consistency between features and introduce a novel notion of actionability that can cover various constraints and prioritize different preferences. Our main contributions are as follow:

- We propose a modular explanation framework that handles the model- and user/domain-level desiderata in a consecutive and structured manner.
- We devise novel and efficient solutions for every requirement that are formulated in a multi-objective optimization framework.
- We introduce a model-agnostic approach to generate multiple, diverse explanations for any black-box model in tabular classification and regression settings.
- We demonstrate the importance of various requirements in counterfactual generation and the efficacy of our modular framework in addressing them through extensive validation and benchmark experiments.
- We provide a multi-purpose and flexible benchmark for the research community: https://github.com/peymanrasouli/CARE.

## 2 THE FRAMEWORK OF CARE

State-of-the-art works take into account a subset of the stated properties for counterfactual generation. However, as we later demonstrate by an illustrative example, every property plays a unique and crucial role in creating feasible and actionable explanations. We propose a modular framework that formulates the necessary properties in a consecutive and structured manner. Figure 1 illustrates the overall framework of CARE. It consists of four modules each fulfills some specific properties for actionable recourse including VALIDITY, SOUNDNESS, COHERENCY, and ACTIONABILITY. Low-level modules mostly contain model-related objectives, while high-level modules target user/domain-related goals. VALIDITY module acts as a basis for other modules, and counterfactuals can be generated regardless of the presence of other intermediate modules. In other words, it is possible to disable/enable desired modules in the hierarchy. Using this structure, we can observe the effect of the different properties on the generated counterfactuals.

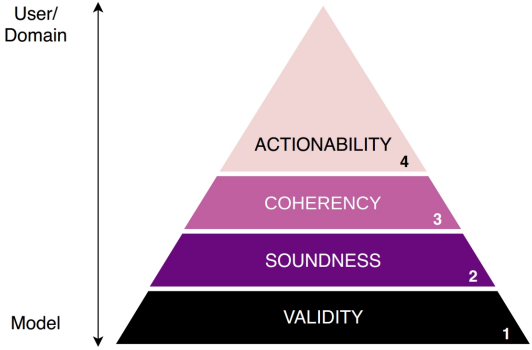

**Figure 1: CARE's framework: modular hierarchy.**

To provide an insight on how the different properties affect the generated counterfactual explanations, we explain an instance of the *Adult Income* data set [6], denoted by $X_{original}$, using incremental configurations of CARE's modules. The results are reported in Table 1. It can be seen that the valid counterfactual ($E_{valid}$) is a minimally modified variant of the original input that only provides the desired outcome irrespective of feasibility and actionability of changes. In contrast, the sound counterfactual ($E_{sound}$) lies on the data manifold and appears more realistic, however, it has not established the correlation between features perfectly (the values of features **relationship** and **sex** do not conform with each other). The coherent counterfactual ($E_{coherent}$) resolves the consistency shortcoming of sound counterfactuals and creates a coherent state for input features (features **relationship** and **sex** are consistent). Nevertheless, a statistically sound and coherent counterfactual may not be actionable from the user's perspective (changing sensitive features like **sex** is impractical). Hence, an actionable explanation ($E_{actionable}$) takes into account the user's preferences containing the name of mutable/immutable features, possible values, and their importance to create a realistic and actionable recourse.

The stated example demonstrates the importance of every property as well as the exclusive role of CARE's modules in handling these properties. Our framework can be used as a benchmark for explanation methods sharing similar properties, as it allows proper comparison by enabling specific modules in the hierarchy. CARE has two main phases: *fitting* and *explaining*; the former creates an *explainer* based on the training data and the ML model, while the latter generates counterfactuals for every input instance using the created *explainer*. In the following, we describe the CARE's modules and the optimization algorithm for generating explanations.

## 2.1 Valid Counterfactual Explanations

We define a valid counterfactual explanation as an instance similar to the input with minimum changes to features that results in the desired outcome. Consider $(X^m, Y)$ as a data set, where $X^m$ is an $m$ dimensional feature space and $Y$ is a target set. Let $f$ be a black-box model trained on $(X^m, Y)$ that map an input to a target, i.e., $f : X^m \rightarrow Y$. Targets can be either discrete classes or continuous values depending on the prediction task (classification or regression). For a given input $x$ with $f(x) = y$ and a desired outcome $y'$, our goal is to find a counterfactual $x'$ as close to $x$ as possible such that $f(x') = y'$. VALIDITY module satisfies three main objectives: desired outcome, minimum feature distance, and maximum sparsity. Maximum sparsity enhances the interpretability of the generated counterfactual explanations and is inline with minimum feature distance. We define three cost functions for the stated objectives that are minimized during the optimization process.

### 2.1.1 Desired Outcome.
We evaluate the prediction of $f$ for a generated counterfactual $x'$ with respect to the desired outcome. Accessing the black-box prediction function allows us to define different measurements for the desired outcome. For the classification task, we use the Hinge loss function:

$$O_{outcome}^C(x', c, p) = \max(0, (p - f_c(x'))) \quad (1)$$

where $f_c(x')$ is the prediction probability of $x'$ for the desired class $c$ and $p$ is a probability threshold that leads to a counterfactual with a desired level of confidence. This cost function considers all counterfactuals above the threshold $p$ as valid counterfactuals. The desired outcome for the regression task is evaluated as follows:

$$O_{outcome}^R(x', r) = \begin{cases} 0, & \text{if } f(x') \in r \\ \min_{r' \in r} |f(x') - r'|, & \text{otherwise} \end{cases} \quad (2)$$

where $f(x')$ is the predicted response for the counterfactual and $r = [lb, ub]$ is a desired response range. The devised cost function considers any predicted response within the range $r$ as valid (zero cost), otherwise the absolute distance between the prediction and the closest bound ($lb$ or $ub$) is considered as the cost of $x'$.

### 2.1.2 Feature Distance.
We employ Gower distance [8] to calculate the distance between features in a mixed-feature setting (i.e., data set contains both categorical and numerical features). Given an original input $x$ and a counterfactual instance $x'$, we measure their feature value distance by summation over the difference between feature values:

$$O_{distance}(x, x') = \frac{1}{m} \sum_{j=1}^{m} \delta(x_j, x'_j) \quad (3)$$

where $m$ is the number of features, and $\delta$ is a metric function that returns a distance value depending on the type of the feature:

$$\delta(x_j, x'_j) = \begin{cases} \dfrac{1}{R_j} |x_j - x'_j|, & \text{if } x_j \text{ is numerical} \\ \mathbb{1}_{x_j \neq x'_j}, & \text{if } x_j \text{ is categorical} \end{cases} \quad (4)$$

where $R_j$ is the value range of feature $j$ that is extracted from the training data, and $\mathbb{1}$ is a binary indicator function.

### 2.1.3 Sparsity.
To have a highly interpretable explanation, a counterfactual should alter a minimum number of features. Minimum feature distance is not equivalent to the minimum number of changed features. Therefore, we define the cost function $O_{sparsity}$ for counting the number of altered features:

$$O_{sparsity}(x, x') = \sum_{j=1}^{m} \mathbb{1}_{x_j \neq x'_j} \quad (5)$$

**Table 1: Counterfactual explanations generated using incremental configurations of CARE's modules for an instance from the *Adult Income* data set.**

| Instance | age | capital-gain | capital-loss | hours-per-week | work-class | education | marital-status | occupation | relationship | race | sex | native-country | Class |
|---|---|---|---|---|---|---|---|---|---|---|---|---|---|
| $X_{original}$ | 35 | 0 | 0 | 40 | Private | HS-grad | Divorced | Sales | Unmarried | White | Female | United-States | ≤50K |
| $E_{valid}$ | – | 7688 | – | – | – | – | – | – | – | – | – | – | >50K |
| $E_{sound}$ | 36 | – | – | 50 | – | – | Married | Exec-managerial | Husband | – | – | – | >50K |
| $E_{coherent}$ | 38 | – | – | 60 | – | – | Married | Transport-moving | Husband | – | Male | – | >50K |
| $E_{actionable}$ | 42 | – | – | 45 | – | Masters | – | Prof-specialty | – | – | – | – | >50K |

where $m$ is the number of features in the data set. This function penalizes every change in features regardless of their type.

## 2.2 Sound Counterfactual Explanations

As we mentioned earlier, a sound counterfactual should originate from the observed data (*proximity*) and connect to the existing knowledge (*connectedness*). Meeting these two conditions results in an inlier instance located in a region in the decision surface where the model has a high level of confidence. Therefore, our SOUNDNESS module has two fitness functions that are maximized during the optimization process.

### 2.2.1 Proximity. 
Proximity indicates that the counterfactual instance lies in the neighborhood of the ground-truth samples that are predicted correctly by the model and have the same target value as the counterfactual. We utilize the proximity evaluation metric introduced in [12] as an objective function for counterfactual generation. Let $x$ be our original input and $x'$ be a generated counterfactual. We refer to $X^{f(x')}$ as the set of instances in the data set that are predicted correctly by $f$ and belong to the same class (in classification task) or response range (in regression task) as $x'$. Consider $a_0 \in X^{f(x')}$ is the closest instance to $x'$, i.e., $a_0 = \arg\min_{a_i \in X^{f(x')}} D(x', a_i)$, where $D$ is a distance metric (e.g., Minkowski). The counterfactual $x'$ fulfills the proximity criterion if it has the same distance to $a_0$ as $a_0$ has to the rest of the data ($X^{f(x')}$). The formal definition of proximity is as follows:

$$proximity(x') = \frac{D(x', a_0)}{\min_{a_i \in X^{f(x')}} D(a_0, a_i)} \quad (6)$$

A lower value of $proximity(x')$ refers to an inlier counterfactual that is located at a reasonable distance from the training data that are predicted identically. According to [1], the formal definition of *proximity* (Eq. 6) corresponds to the Local Outlier Factor (LOF) with a neighborhood size of $K = 1$, which is a well-known model for outlier detection. In the *fitting* phase of the explainer, we create an LOF model for every class/response range of the samples in the training data that are predicted correctly by the model $f$. During the *explaining* phase and via the objective function $O_{proximity}$, we invoke the LOF model related to the class/response range of the counterfactual $x'$ to identify its status; if $x'$ is an inlier, the model outputs 1, otherwise, it returns 0 that refers to an outlier. Thus, the goal is to maximize $O_{proximity}$ for every counterfactual instance.

### 2.2.2 Connectedness. 
Connectedness implies the counterfactual instance is the result of existing knowledge and not a consequence of an artifact of the ML model. Such a counterfactual is continuously connected to the observed data (knowledge) using a topological notion of path. Along this path, features change smoothly and coherently, and each instance in the path is correlated with the preceding and succeeding instances. This property is thus complementary to proximity, in which the counterfactual is close to a real data point but is not necessarily linked to the majority of the data. Similarly, we benefit from the connectedness evaluation metric proposed in [12] as an objective function for counterfactual generation. The continuous path can be approximated by the notion of $\epsilon$-*chainability* (with $\epsilon > 0$) between two instances $e$ and $a$, meaning that a finite sequence $X_N = e_0, e_1, .., e_N$, where $X_N \subset X$, exists such that $e_0 = e$, $e_N = a$, and $\forall i < N, D(e_i, e_{i+1}) < \epsilon$. Let $x'$ be a counterfactual instance for an input $x$. We say counterfactual $x'$ is $\epsilon$-connected to $a \in X$ if $f(x') = f(a)$ and there exist an $\epsilon$-chain $X_N$ between $x'$ and $a$ such that $\forall e \in X_N, f(e) = f(x')$.

Although assessing $\epsilon$-connectedness seems complex, its definition resembles the DBSCAN clustering algorithm [7]. We can acknowledge that $x'$ is $\epsilon$-connected to $a \in X$, if $x'$ and $a$ belong to the same cluster of DBSCAN algorithm with parameters epsilon= $\epsilon$ (maximum distance between two samples) and min_samples= 2 (number of samples in a neighborhood). Using DBSCAN clustering for every counterfactual instance is not computationally efficient. Moreover, finding an optimal epsilon parameter, which highly impacts the clustering results, for every class/response range is challenging. To remedy the stated issues, we employ a generalized version of DBSCAN, called HDBSCAN [2]. This algorithm adaptively selects the best epsilon value to adaptively produce stable clusters. We avoid computational complexity by creating a HDBSCAN model on a large amount of samples and then querying the model for predicting the cluster of a potential counterfactual instance. Since one sample is not likely to alter the shape of the created clusters for the ground-truth data, we achieve a fairly accurate measurement of *connectedness*. Moreover, it does not require updating the clustering model, making the assessment procedure computationally efficient.

We define the objective function $O_{connectedness}$ to connect the generated counterfactuals to the existing knowledge. We categorize the ground-truth samples w.r.t every class/response range that are predicted correctly by the model $f$. A clustering model for every category is then constructed within the *fitting* phase of the explainer. In the *explaining* phase and using objective function $O_{connectedness}$, the clustering model corresponding to the class/response range of the counterfactual $x'$ is queried. If $x'$ is assigned to a cluster, the function returns 1, otherwise, it returns 0, which indicates $x'$ is not

---

**Algorithm 1** Correlation Models

---

**Input:** observed data $X$, correlation threshold $\rho$, score threshold $\tau$, number of features $m$, type of features $F$

**Output:** correlation models $\mathcal{M}$

1: Initialize $\mathcal{M} = \{\}$
2: $corr = \text{CalculateCorrelations}(X, F, \rho)$
3: $X_{train}, X_{val} = \text{SplitData}(X, train\_percent = 0.8)$
4: **for** $j = 1$ **to** $m$ **do**
5:     $inputs = \text{FindCorrelatedFeatures}(corr, j)$
6:     $X_{train}^j, Y_{train}^j = X_{train}[inputs], X_{train}[j]$
7:     $X_{val}^j, Y_{val}^j = X_{val}[inputs], X_{val}[j]$
8:     $model = \text{ConstructModel}(X_{train}^j, Y_{train}^j)$
9:     $score = \text{ValidateModel}(model, X_{val}^j, Y_{val}^j)$
10:     **if** $score \geq \tau$ **then**
11:        $\mathcal{M}_{inputs}^j, \mathcal{M}_{model}^j, \mathcal{M}_{score}^j = inputs, model, score$
12:     **end if**
13: **end for**

---

connected to the known knowledge. Hence, the goal is to maximize $O_{connectedness}$ for every counterfactual instance.

### 2.3 Coherent Counterfactual Explanations

Preserving the relationship between changed/unchanged features is essential for creating a coherent and actionable counterfactual explanation. Fulfilling this property is even more important when the output is a personalized, actionable recourse. By knowing the relationship of the domain's features, one can formulate unary and binary constraints in simple terms. For example, "age does not decrease" or "education level increment causes age increment". However, relationships over multiple features can lead to complex constraints that is hard to specify manually.

We introduce a novel approach to impose high-order correlation constraints over multiple features. Using correlation information extracted from the training data, we construct predictive models (also referred as correlation models) that are exploited to preserve the coherency of counterfactual features. This technique is suitable when the input features are unknown, and knowledge about the domain is limited. The COHERENCY module uses a two-phase approach: 1) constructing correlation models in the *fitting* phase (Algorithm 1), and 2) exploiting the created models in the objective function $O_{coherency}$ during the *explaining* phase (Algorithm 2).

Algorithm 1 starts by extracting feature correlations from the training data $X$ using Pearson's R, Correlation Ratio, and Cramer's V for pairs of numerical-numerical, numerical-categorical, and categorical-categorical features, respectively [3]. This creates a symmetric correlation matrix with scaled values $corr^{m \times m} \in [0, 1]$. We consider every feature $j$, $j \in \{1..m\}$, as a dependent variable that is predicted by its correlated features (independent variables) denoted by $inputs$ (Line 5). Our goal is to create computationally efficient classification and regression models (e.g., CART [14] and Ridge [16]) for every categorical and numerical features, respectively. Features are considered correlated if they have a correlation value above threshold $\rho \in [0, 1]$. We only consider reliable models that have a predictive score (F1-score or $R^2$-score) above threshold

$\tau \in [0, 1]$ (Line 10). This hyper-parameter has a major role in the overall performance of coherency-preservation and determines the magnitude of correlation constraints. At the end (Line 11), if there exist a reliable model for a feature $j$, $j \in \{1..m\}$, the correlation model $\mathcal{M}^j$ will include a triple $\{inputs, model, score\}$ containing the inputs of the model, trained model, and score of the model for feature $j$.

---

**Algorithm 2** Coherency Objective ($O_{coherency}$)

---

**Input:** input $x$, counterfactual $x'$, correlation models $\mathcal{M}$

**Output:** coherency cost $\xi$

1: Initialize $\xi = 0$
2: $L = \text{ChangedFeatures}(x, x')$
3: **for all** $j \in L$ **do**
4:     **if** $j \in \mathcal{M}$ **then**
5:        $inputs, model, score = \mathcal{M}_{inputs}^j, \mathcal{M}_{model}^j, \mathcal{M}_{score}^j$
6:        $x'_{temp} = x'$
7:        $x'_{temp}[j] = model(x'[inputs])$
8:        $distance = \delta(x', x'_{temp})$
9:        $cost = score * distance$
10:        $\xi = \xi + cost$
11:     **end if**
12: **end for**

---

In Algorithm 2, the created models $\mathcal{M}$ are used to establish the coherency among the counterfactual's features. First we identify the changed features in the counterfactual $x'$, denoted by $L$. For a feature $j \in L$, if correlation model $\mathcal{M}^j$ exists, it is used to predict the value of the feature, creating a temporary counterfactual $x'_{temp}$ (i.e., a copy of $x'$ with the predicted value for feature $j$). The model $\mathcal{M}^j$ makes prediction based on the values of the correlated features $inputs$ (Line 7). Therefore, it takes into account complex relationships between multiple features. We measure the preserved coherency by calculating the distance between the counterfactual $x'$ and the temporary counterfactual $x'_{temp}$ (Line 8). The intuition is that if the values of correlated features conform with each other, the prediction of the model for feature $j$ will be close to the counterfactual's value for the feature, i.e., $x'_{temp}[j] \simeq x'[j]$, leading to a low $distance$ between $x'$ and $x'_{temp}$. Although we filtered out unreliable models in Algorithm 1, here, we weigh the $distance$ according to the $score$ of the correlation model to have a truthful measurement of the coherency cost (Line 9). This procedure is repeated for every feature, and the output of $O_{coherency}$ is the total coherency cost $\xi$ that is minimized during the optimization process.

### 2.4 Actionable Recourse

Although previous modules provide a sound and coherent counterfactual explanation, it may not be necessarily actionable from the user's perspective (for example, recommending a change to **race** or **gender**). The objective of ACTIONABILITY module (Figure 1) is to allow users to define preferences over features to guarantee the actionability of a recourse. We propose a constraint language (outlined in Table 2) to provide the user with a flexible set of constraints for features. The language provides diverse operators for numerical and categorical features. We also present the notion of *constraint importance* to weigh the constraints according to their importance for the user. For example, two constraints "fix the race" and "constrain balance between [5000$, 10000$]" obviously have different importances, as exceeding the balance range may be acceptable, but changing race is not tolerable. Without considering constraint importance, the optimization algorithm makes no difference between a set of constraints. By weighing constraints, we can prioritize them, and therefore, the optimization algorithm avoids to overstep the highly important ones. Thus, we define the user's preference $\mathcal{P}$ as a set of constraint triples in the form of $C_j = (feature_j, constraint_j, importance_j)$ for a desired feature $j$, $j \in \{1..m\}$, where $m$ is the total number of features, i.e., $\mathcal{P} = \{C_x, C_y, .., C_z\}, x, y, z \in \{1..m\} \land x \neq y \neq z$. An example preference can be $\hat{\mathcal{P}} = \{(age, ge, 4), (race, fix, 10)\}$. Algorithm 3 outlines our proposed objective function $O_{actionability}$ which computes the actionability cost $\eta$ for a particular counterfactual $x'$ according to the user's preference $\mathcal{P}$. For every constraint triple $\forall C \in \mathcal{P}$, Algorithm 3 checks the satisfiability of $C_{constraint}$; if $C_{constraint}$ is not satisfied, then $C_{importance}$ is added to the actionability cost $\eta$. The output of $O_{actionability}$ is the total incurred actionability cost $\eta$ that is minimized during the optimization process.

**Table 2: Constraint language.**

| Feature Type | Constraint | Description |
|---|---|---|
| Numerical | $fix$ | fix the current value |
| | $l$ | greater than the current value |
| | $g$ | less than the current value |
| | $le$ | less than or equal to the current value |
| | $ge$ | greater than or equal to the current value |
| | $[lb, ub]$ | a range of numerical values |
| Categorical | $fix$ | fix the current value |
| | $\{v_1, .., v_n\}$ | a set of categorical values |

**Table 3: Summary of information of the used data sets.**

| Data set | # Samples | # Numerical | # Categorical | Task |
|---|---|---|---|---|
| *Adult Income* | 48842 | 6 | 8 | Classification |
| *COMPAS* | 7214 | 4 | 7 | Classification |
| *Default of Credit Card Clients* | 30000 | 20 | 3 | Classification |
| *Boston House Prices* | 506 | 12 | 1 | Regression |

---

**Algorithm 3** Actionability Objective ($O_{actionability}$)

---

**Input:** input $x$, counterfactual $x'$, user's preference $\mathcal{P}$
**Output:** actionability cost $\eta$
1: Initialize $\eta = 0$
2: **for all** $C \in \mathcal{P}$ **do**
3:     $F, C, I = C_{feature}, C_{constraint}, C_{importance}$
4:     $satisfied = $ CheckSatisfiability$(C, x[F], x'[F])$
5:     **if** $satisfied = $ False **then**
6:         $\eta = \eta + I$
7:     **end if**
8: **end for**

---

### 2.5 Multi-objective Optimization Framework

In this section, we adopt Non-dominated Sorting Genetic Algorithm III (NSGA-III) [5] to solve our multi-objective optimization problem. Compared to other evolutionary algorithms, NSGA-III performs well with differently scaled objective values and generates diverse solutions. The first property is essential for a multi-objective counterfactual explanation method where there exists a combination of fitness and cost functions with different ranges of output and conflict goals. The second property is useful in the sense of actionability, as providing a diverse set of solutions increases the chance of obtaining a recourse complying with the user's circumstance. The objective set $Obj$ defines the CARE's hierarchy:

$$Obj = \Big\{ \{\downarrow O_{outcome}, \downarrow O_{distance}, \downarrow O_{sparsity}\}_1,$$
$$\{\uparrow O_{proximity}, \uparrow O_{connectedness}\}_2, \qquad (7)$$
$$\{\downarrow O_{coherency}\}_3, \{\downarrow O_{actionability}\}_4 \Big\}$$

Every set in $Obj$ corresponds to a module annotated by a subscript number, i.e., **1, 2, 3,** and **4** (also shown in Figure 1). Arrows indicate the type of objectives, either cost or fitness function, which should be minimized or maximized, respectively. As we mentioned earlier, the first set (VALIDITY module) is always present, while other sets (modules) can be arbitrarily included. For example, we may create the configuration $\{1, 2\}$ which only contains the VALIDITY and SOUNDNESS modules. This modular structure allows our method to be used for both counterfactual explanation (configuration $\{1, 2\}$) and actionable recourse generation (configuration $\{1, 2, 3, 4\}$). Since the defined objective functions make no assumption about the nature and internal structure of the prediction model $f$, CARE is applicable on any ML model created for tabular classification and regression tasks. Moreover, the NSGA-III algorithm can automatically handle mixed-feature data dispensable of auxiliary operations (such as one-hot encoding and imposing hard-constraints), making CARE suitable for mixed-feature tabular data sets.

## 3 EMPIRICAL EVALUATIONS

In this section, first, we describe the evaluation setup including data sets, models, and hyper-parameters. Second, we validate the importance of the stated desiderata and the effectiveness of CARE's modules in handling them using multiple experiments. Finally, we demonstrate the overall performance of CARE by benchmarking against two state-of-the-art explanation methods.

### 3.1 Evaluation Setup

We mainly use standard data sets from the UCI Machine Learning Repository [6], except the *COMPAS* data set that can be found at [19]. Summary of information of the used data sets is reported in Table 3. The numerical features were standardized by removing the mean and scaled to unit variance. We converted original categorical features to ordinal encoding and used their corresponding one-hot encoding for creating black-box models.

We split the data sets into 80% *train set* and 20% *test set*. The *train set* was used for creating black-box and the explainer models while the *test set* was used for generating explanations. We created a Multilayer Perceptron Neural Networks (**NN**) consisted of two hidden layers each with 50 neurons and a Gradient Boosting Machines (**GB**) comprised of 100 estimators as black-box classifiers and regressors. Table 4 reports the performance of created models on *test set* in terms of F1-score (classification task) and $R^2$-score (regression task).

**Table 4: Performance of the black-box models.**

| Data set | NN | GB |
|---|---|---|
| *Adult Income* | 0.849 | 0.860 |
| *COMPAS* | 0.788 | 0.808 |
| *Default of Credit Card Clients* | 0.794 | 0.797 |
| *Boston House Prices* | 0.829 | 0.916 |

We used 4-quantiles as cutting points for intervals in the regression task and generated counterfactuals from the neighbor interval for every input. In addition, for the classification task, we set the probability threshold as $p = 0.5$ and generated counterfactuals from the opposite class. We used CART decision trees [20] for categorical features and Ridge regression models [16] for numerical features as correlation models (refer to Algorithm 1). We set the minimum correlation threshold as $\rho = 0.1$. To adjust the threshold of the model's score (i.e., $\tau$) with respect to each data set, we set $\tau$ as the median value of all models' scores and filtered out the correlation models that have a score below $\tau$. Accordingly, we achieved $\tau = 0.72$, $\tau = 0.59$, $\tau = 0.84$, and $\tau = 0.64$ for *Adult Income, COMPAS, Default of Credit Card Clients*, and *Boston House Prices* data sets, respectively. For NSGA-III optimization algorithm, we used the two-point crossover operator with percentage $pc = 60\%$ and the polynomial mutation operator with percentage $pm = 30\%$. The number of generations was set to $n_{generation} = 10$. Although it is possible to define arbitrary population size $n_{population}$, we determined it adaptively with respect to the number of objectives using a standard formula described in the original paper of NSGA-III [5]. This approach is more suitable for our modular framework in which the number of objectives would vary with respect to different configurations. Regarding user preferences for actionable recourse, we only set global constraints according to our basic knowledge about the domains for all inputs. For example, constraint *fix* (fix the current value) was set for features like **race** and **sex** while constraint *ge* (greater than or equal to the current value) was set for feature **age**. Thus, actionable recourses are not locally personalized with respect to every specific input. We set equal importance values for all constraints, i.e., $importance = 1.0$, to have a fair comparison regarding the actionability objective function (i.e., $O_{actionability}$) among baseline methods. CARE has been developed using Python programming language, and experiments were run on a system with Intel Core i7-8650U processor and 32GB of memory. A complete implementation of the method, including validation and benchmark experiments, is available at: https://github.com/peymanrasouli/CARE.

## 3.2 Validation of Soundness Module

We visualize the impact of SOUNDNESS module in generating sound and realistic counterfactuals on *Iris* and *Moon* data sets [6]. We created Gradient Boosting (**GB**) classifiers on these data sets and generated $cf_V$ and $cf_S$ counterfactuals using configurations $\{1\}$ and $\{1, 2\}$, respectively. Figure 2(a) demonstrates counterfactuals from the furthest class for a sample from *Iris* data set, and Figure 2(b) depicts counterfactuals from the opposite class for an instance from *Moon* data set. The counterfcatuals are annotated with the values of $O_{proximity}$ and $O_{connectedness}$ denoted by $p$ and $c$, respectively.

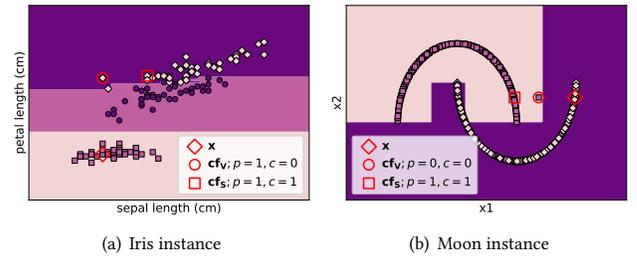

(a) Iris instance     (b) Moon instance

**Figure 2: Soundness validation results.**

By observing the location of the generated counterfactuals in Figures 2(a) and 2(b), the importance of SOUNDNESS module is revealed. It can be seen that minimum distance is the only important criterion for $cf_V$ counterfactuals while $cf_S$ counterfactuals are associated to the same-class training data ($p = 1$ and $c = 1$), therefore they are connected to the previous knowledge and achieve a high prediction probability from the classifier. We stated that *proximity* is different than *connectedness* and they are complementary criteria for a sound counterfactual explanation. Figure 2(a) demonstrates this distinction clearly as $cf_V$ is located at the proximity of the training data ($p = 1$), but it is not connected to a high density region ($c = 0$) which declines its interpretability and actionability.

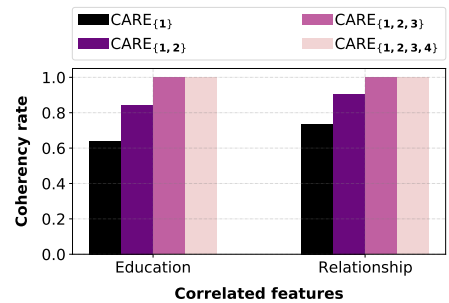

**Figure 3: Evaluation of coherency-preservation.**

## 3.3 Validation of Coherency Module

To validate the impact of COHERENCY module, we have conducted an experiment using *Adult* data set [6]. In this data set, categorical features **education** and **education-num** are fully correlated to each other, as there is a one-to-one correspondence between their values. Moreover, there is a strong relationship among features **marital-status**, **relationship**, and **sex**. We refer to the former set of correlated features as "Education" and the latter as "Relationship." A coherent counterfactual explanation should change all features in a set accordingly; otherwise, it creates an inconsistent feature state. We define the *coherency rate* as the normalized number of counterfactuals in which the consistency between features in every correlation set ("Education" and "Relationship") is preserved.

**Table 5: Effectiveness of CARE's modules.**

| Data set | Configuration | $\downarrow O_{outcome}$ | $\downarrow O_{distance}$ | $\downarrow O_{sparsity}$ | $\uparrow O_{proximity}$ | $\uparrow O_{connectedness}$ | $\downarrow O_{coherency}$ | $\downarrow O_{actionability}$ |
|---|---|---|---|---|---|---|---|---|
| Adult Income | {1} | 0.00 ± 0.0 | **0.04 ± 0.0** | **1.72 ± 1.0** | 0.63 ± 0.5 | 0.29 ± 0.5 | 0.14 ± 0.3 | 0.09 ± 0.3 |
| | {1, 2} | 0.00 ± 0.0 | 0.10 ± 0.1 | 3.04 ± 1.8 | 0.97 ± 0.2 | 0.96 ± 0.2 | 0.18 ± 0.4 | 0.35 ± 0.6 |
| | {1, 2, 3} | 0.00 ± 0.0 | 0.10 ± 0.1 | 2.98 ± 1.8 | 0.98 ± 0.1 | **0.97 ± 0.2** | 0.00 ± 0.1 | 0.31 ± 0.5 |
| | {1, 2, 3, 4} | **0.00 ± 0.0** | 0.10 ± 0.1 | 3.03 ± 1.9 | **0.98 ± 0.1** | 0.95 ± 0.2 | **0.00 ± 0.0** | **0.06 ± 0.2** |
| Boston House Prices | {1} | 0.00 ± 0.0 | **0.02 ± 0.0** | **2.25 ± 1.9** | 0.72 ± 0.5 | 0.86 ± 0.3 | 0.06 ± 0.1 | **0.05 ± 0.2** |
| | {1, 2} | 0.00 ± 0.0 | 0.05 ± 0.1 | 3.75 ± 2.8 | 1.00 ± 0.0 | 1.00 ± 0.0 | 0.11 ± 0.2 | 0.18 ± 0.4 |
| | {1, 2, 3} | 0.00 ± 0.0 | 0.07 ± 0.1 | 3.87 ± 3.4 | 1.00 ± 0.0 | 0.99 ± 0.1 | 0.02 ± 0.0 | 0.19 ± 0.4 |
| | {1, 2, 3, 4} | **0.00 ± 0.0** | 0.06 ± 0.1 | 3.44 ± 3.1 | **1.00 ± 0.0** | **1.00 ± 0.0** | **0.02 ± 0.0** | 0.10 ± 0.3 |

Figure 3 depicts the results of the coherency validation experiment. We created an **NN** model using the *train set* and generated a set of ten counterfactual explanations for 500 samples form the *test set* using different configurations of CARE. According to Figure 3, a sound explanation method (CARE$_{\{1,2\}}$) considerably resolves the inconsistency existing in valid explanations (CARE$_{\{1\}}$), however, it does not guarantee generating coherent explanations for all inputs. On the contrary, configurations equipped with the COHERENCY module (i.e., CARE$_{\{1,2,3\}}$ and CARE$_{\{1,2,3,4\}}$) substantially preserve the correlation among features, leading to maximum *coherency rate*. The efficiency of our algorithm in handling complex relationships will be validated using a user study in future work.

### 3.4 Effectiveness of CARE's Modules

Modules in the CARE's hierarchy are rigorously designed to handle important desiderata for the counterfactual generation. Earlier, we demonstrated the role of CARE's modules in approaching different properties via an illustrative example (Table 1). In this section, we quantitatively evaluate the impact of each module by conducting the following experiment. We consider four configurations {1}, {1, 2}, {1, 2, 3}, and {1, 2, 3, 4}. We are interested to know the behavior of CARE when some modules are absent. For example, how much a counterfactual generated by configuration {1} will satisfy objectives defined in configuration {1, 2}. Using this information we can determine the importance of the devised modules. To this end, we created **GB** models using the *train set* of *Adult Income* and *Boston House Prices* data sets [6] and generated counterfactual explanations using the specified module combinations for 500 and 100 samples of their *test set*, respectively. Eventually, their results regarding the objectives defined in the last configuration (i.e., {1, 2, 3, 4}), which contains all CARE's modules, were measured. By observing the results demonstrated in Table 5, we can conclude several points about the performance of different modules:

- CARE generates valid counterfactuals for all configurations and data sets since validity is the basis of our methodology. These counterfactuals are best at fulfilling the $O_{distance}$ and $O_{sparsity}$ objectives because closeness and sparsity are the only essential goals in this setting. Moreover, their cost for the $O_{actionability}$ objective is usually low since changing immutable features leads to a counterfactual with a dramatic distance to the original input, therefore, the optimization algorithm usually avoids manipulating such features.

- The SOUNDNESS module has significantly improved the $O_{proximity}$ and $O_{connectedness}$ objectives in configurations {1, 2}, {1, 2, 3}, and {1, 2, 3, 4}. Meanwhile, since sound counterfactuals originate from high-density regions, their $O_{distance}$ and $O_{sparsity}$ costs are expectedly increased.

- The devised COHERENCY module has efficiently preserved the relationship among counterfactual features, resulting in $O_{coherency} \approx 0$ for configurations equipped with the module (i.e., {1, 2, 3} and {1, 2, 3, 4}). This shows the generalization of our method for high dimensional data sets where there exist complex correlations between features.

- By comparing the results of two configurations {1, 2, 3} and {1, 2, 3, 4} we realize that given a sound and coherent counterfactual, generating an actionable recourse does not negatively influence the other objectives. CARE creates actionable recourse based on sound and coherent counterfactuals. Though, for some inputs and their corresponding constraints, it is impossible to find an actionable recourse that fulfills the objectives in the precedent modules (modules 2 and 3), causing $O_{actionability}$ cost.

### 3.5 Benchmark Results

To evaluate the performance of CARE, we compared it with state-of-the-art explanation methods CFPrototype [23] and DiCE [17]. We chose CFPrototype because it uses a loss function called *prototype loss* to generate interpretable counterfactuals located in the proximity of the same-class training data. We selected DiCE due to its diverse counterfactual generation property and its ability to impose actionability constraints on the input features. To balance between diversity and proximity of the generated counterfactuals in DiCE, we set the corresponding hyper-parameters as $\lambda_1 = 1.0$ and $\lambda_2 = 1.0$. We used CFPrototype with the default hyper-parameters stated in the paper. The methods were applied on three classification data sets *Adult Income*, *COMPAS* [19], and *Default of Credit Card Clients* (*Default of CCC* for short) [6]. We created an **NN** black-box model for every data set using their *train set* and explained 500 samples from their *test set*. The number of generated counterfactuals in CARE and DiCE was set to $N = 10$.

*3.5.1 Performance Evaluation.* Table 6 reports the evaluation of counterfactuals with respect to the CARE's objective functions. CARE's counterfactuals originate from the distribution of the same-class training data and are connected to the existing knowledge by a continuous path; therefore, it achieved the highest values for $O_{proximity}$ and $O_{connectedness}$ fitness functions. Moreover, the coherency in CARE's explanations is fully respected in both data sets (zero cost for $O_{coherency}$). It is noteworthy that the amount of coherency-preservation is directly related to the number of correlation models that we had generated; the more reliable correlation models exist, the more complex correlations are preserved. As we explained in Section 3.4, we can effortlessly generate an actionable recourse if we only consider its distance to the original input. But, CARE is designed to follow the order of the modules in its hierarchy and prioritize sound and coherent explanations. For some inputs and their corresponding constraints, it is impossible to meet the two mentioned properties, resulting in non-actionable recourses (non-zero cost for $O_{actionability}$). Last but not least, CARE generates valid counterfactuals for every input (zero cost for $O_{outcome}$).

**Table 6: Performance evaluation using objective functions: CARE vs. CFPrototype and DiCE.**

| Data set | Method | ↓ $O_{outcome}$ | ↓ $O_{distance}$ | ↓ $O_{sparsity}$ | ↑ $O_{proximity}$ | ↑ $O_{connectedness}$ | ↓ $O_{coherency}$ | ↓ $O_{actionability}$ |
|---|---|---|---|---|---|---|---|---|
| *Adult Income* | DiCE | 0.17 ± 0.2 | 0.27 ± 0.1 | 5.77 ± 1.4 | 0.19 ± 0.4 | 0.01 ± 0.1 | 1.85 ± 0.6 | **0.00 ± 0.0** |
| | CFPrototype | 0.01 ± 0.0 | **0.08 ± 0.1** | **2.88 ± 1.3** | 0.60 ± 0.5 | 0.14 ± 0.3 | 0.52 ± 0.6 | 0.24 ± 0.5 |
| | CARE | **0.00 ± 0.0** | 0.11 ± 0.1 | 2.97 ± 1.9 | **0.97 ± 0.2** | **0.95 ± 0.2** | **0.00 ± 0.0** | 0.06 ± 0.2 |
| *COMPAS* | DiCE | 0.07 ± 0.2 | 0.38 ± 0.1 | 6.72 ± 1.4 | 0.29 ± 0.5 | 0.04 ± 0.2 | 1.25 ± 0.6 | **0.00 ± 0.0** |
| | CFPrototype | 0.00 ± 0.0 | **0.09 ± 0.1** | **3.22 ± 1.3** | 0.63 ± 0.5 | 0.36 ± 0.5 | 0.10 ± 0.3 | 0.09 ± 0.3 |
| | CARE | **0.00 ± 0.0** | 0.12 ± 0.1 | 3.44 ± 1.9 | **1.00 ± 0.0** | **1.00 ± 0.0** | **0.00 ± 0.0** | 0.16 ± 0.4 |
| *Default of Credit Card Clients* | DiCE | 0.08 ± 0.1 | 0.16 ± 0.0 | 15.42 ± 1.8 | 0.30 ± 0.5 | 0.07 ± 0.3 | 2.19 ± 0.9 | **0.00 ± 0.0** |
| | CFPrototype | 0.00 ± 0.0 | **0.09 ± 0.0** | 10.51 ± 4.8 | 0.21 ± 0.4 | 0.03 ± 0.2 | 1.04 ± 0.7 | 0.50 ± 0.5 |
| | CARE | **0.00 ± 0.0** | 0.12 ± 0.1 | **6.12 ± 5.7** | **0.98 ± 0.1** | **0.98 ± 0.2** | **0.00 ± 0.0** | 0.12 ± 0.3 |

*3.5.2 Coherency Evaluation.* To justify the effectiveness of the COHERENCY module and to demonstrate the behavior of baseline methods regarding coherency-preservation, we benchmarked CARE against CFPrototype and DiCE on the experiment defined in Section 3.3. Figure 4(a) illustrates the coherency rate of explanations with respect to the correlation sets "Education" and "Relationship." It can be seen that CARE outperforms baseline methods by generating coherent counterfactual explanations for every input. This experiment highlights that the statistically soundness objectives (like *proximity* which is considered by CFPrototype and DiCE methods) do not necessarily guarantee to generate coherent explanations.

*3.5.3 Diversity Evaluation.* We benchmarked CARE versus DiCE regarding the diversity of the generated counterfactuals. A diverse set of counterfactual explanations for a particular input provides the user with more alternatives for obtaining the desired outcome. We measure the diversity with respect to feature variation as follows:

$$d_F = 1.0 - \left( \frac{1}{C_N^2} \sum_{i=1}^{N-1} \sum_{j=i+1}^{N} Jaccard(\mathcal{S}_i, \mathcal{S}_j) \right) \quad (8)$$

where $C$ is combination operator, $N$ is the total number of counterfactuals, and $\mathcal{S}$ is the set of explanations. We calculate the Jaccard index between the feature names in every pair of explanations for calculating $d_F$. In our opinion, this is a representative metric for diversity because we are interested in a set of counterfactuals that provides various options for a user, not only changes in one or a few specific features. For some inputs, a few important features determine the decision, which in this case, recommending different values for the features is acceptable. We propose a fair

metric for value-based diversity as follows:

$$d_V = 1.0 - \left( \frac{1}{C_N^2} \sum_{i=1}^{N-1} \sum_{j=i+1}^{N} \frac{1}{|(\mathcal{S}_i \cap \mathcal{S}_j)|} \sum_{\forall k \in (\mathcal{S}_i \cap \mathcal{S}_j)} \mathbb{I}_{\mathcal{S}_i^k = \mathcal{S}_j^k} \right) \quad (9)$$

where $\mathbb{I}$ is a binary indicator function. Instead of computing the difference between feature values for every pair of counterfactuals in $\mathcal{S}$, which is a biased measurement due to the different number of changed features (i.e., sparsity degree) by every baseline method, we measure the difference between the feature values of common features in every pair of explanations. Figure 4(b) illustrates the results of this evaluation. Although we have not formulated diversity as an objective function, the choice of the optimization algorithm (i.e., NSGA-III) has led to promising outcomes. We observe that CARE tends to generate counterfactuals that are different concerning the feature names (high $d_F$). It also performs solid when changing multiple features is not applicable ($d_V \approx 0.6$).

*3.5.4 Computational Complexity Evaluation.* Computational complexity is an essential matter for a counterfactual explanation method. The overall complexity of the NSGA-III algorithm is $O(MN^2)$ where $M$ is the number of objectives, and $N$ is the population size. CARE does not require creating a new model for explaining every instance; once a CARE's explainer for a black-box model and its corresponding data set is built, it takes $O(MN^2)$ to explain every input. We have used an adaptive mechanism defined in the original paper of NSGA-III [5] for setting the population size. It determines the population size based on the number of objective functions. This approach is suitable for our modular framework in which modules are arbitrarily added/removed. Figure 4(c) shows the average time spent by each algorithm for explaining a single instance. Although the used data sets have different feature dimensionality, CARE generates explanations in a reasonable and similar amount of time compared to the baseline methods. Furthermore, the computational complexity of CARE reduces when fewer modules are employed.

## 4 CONCLUSIONS

In this paper, we presented CARE, a modular explanation framework for generating actionable recourse. We demonstrated that a sound counterfactual instance is located in the neighborhood and continuously linked to the same-class ground-truth data points. An actionable recourse can be constructed based on such counterfactuals while preserving the correlation between changed/unchanged features and satisfying user/domain-specific constraints. Modules in the CARE's hierarchy are rigorously designed to address the stated desiderata. The low-level modules handle fundamental and statistical properties related to the model and observed data, while the high-level ones manage the coherency between features and user preferences. Through several experiments, we demonstrated the efficacy of our devised approach in creating feasible and realistic explanations and its superior performance compared to the baselines. CARE's framework can be viewed as a flexible benchmark for counterfactual and actionable recourse generation techniques sharing similar desiderata. In future work, we will apply our approach to a large-scale, real-world use case and evaluate its performance using a user study.

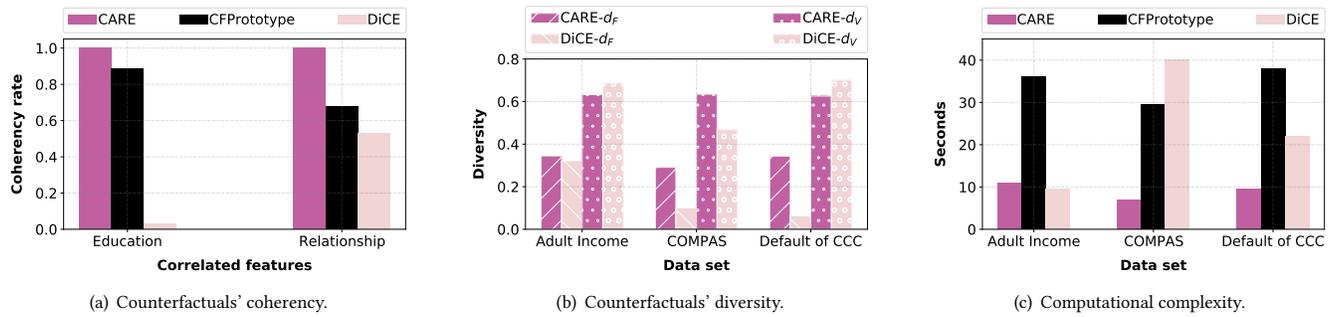

(a) Counterfactuals' coherency.

(b) Counterfactuals' diversity.

(c) Computational complexity.

**Figure 4: Comparison of explanation methods with respect to coherency, diversity, and computational complexity.**